%% file: main.tex
\documentclass[runningheads]{llncs}

\usepackage[mobile]{eccv}

\usepackage{eccvabbrv}

\usepackage{graphicx}
\usepackage{booktabs}

\usepackage[accsupp]{axessibility}  %

\usepackage[pagebackref,breaklinks,colorlinks,citecolor=eccvblue]{hyperref}

\usepackage{orcidlink}

\usepackage{graphicx}
\usepackage{booktabs}
\usepackage{algorithm}
\usepackage{listings}
\usepackage{upquote}  %
\usepackage{xcolor,colortbl}
\lstdefinestyle{overleaf}{
    backgroundcolor=\color[rgb]{0.95,0.95,0.92},   
    commentstyle=\color[rgb]{0,0.6,0},
    keywordstyle=\color{magenta},
    numberstyle=\tiny\color[rgb]{0.5,0.5,0.5},
    stringstyle=\color[rgb]{0.58,0,0.82},
    basicstyle=\ttfamily\footnotesize,
    breakatwhitespace=false,         
    breaklines=true,                 
    captionpos=b,                    
    keepspaces=true,                 
    numbers=left,                    
    numbersep=5pt,                  
    showspaces=false,                
    showstringspaces=false,
    showtabs=false,                  
    tabsize=2
}
\lstdefinestyle{mocov3}{
  backgroundcolor=\color{white},
  basicstyle=\fontsize{7.5pt}{7.5pt}\ttfamily\selectfont,
  columns=fullflexible,
  breaklines=true,
  captionpos=b,
  commentstyle=\fontsize{7.5pt}{7.5pt}\color[rgb]{0.25,0.5,0.5},
  keywordstyle=\fontsize{7.5pt}{7.5pt}\color[rgb]{0.85,0.18,0.50},
}

\begin{document}

\newcommand{\ourmethod}{{\it GS-LRM}}

\newcommand{\secref}[1]{\noindent Sec. \ref{#1}}
\newcommand{\figref}[1]{\noindent Fig. \ref{#1}}
\newcommand{\tabref}[1]{\noindent Tab. \ref{#1}}
\newcommand{\img}{\mathbf{I}}
\newcommand{\tok}{\mathbf{T}}
\newcommand{\boldstart}[1]{\noindent\textbf{#1}}
\newcommand{\boldstartspace}[1]{\medskip\noindent\textbf{#1}}

\title{GS-LRM: Large Reconstruction Model \\
for 3D Gaussian Splatting} 

\titlerunning{GS-LRM: Large Reconstruction Model for 3D Gaussian Splatting}

\author{Kai Zhang$^{*1}$
\quad
Sai Bi$^{*1}$
\quad
Hao Tan$^{*1}$
\quad
Yuanbo Xiangli$^2$ \\
\quad
Nanxuan Zhao$^1$
\quad
Kalyan Sunkavalli$^1$
\quad
Zexiang Xu$^1$\\
}
\institute{$^1$Adobe Research \quad $^2$Cornell University}
\authorrunning{K.\ Zhang, S.\ Bi, H.\ Tan et al.}

\maketitle
\begin{figure}[t]
    \centering
    \includegraphics[width=\textwidth]{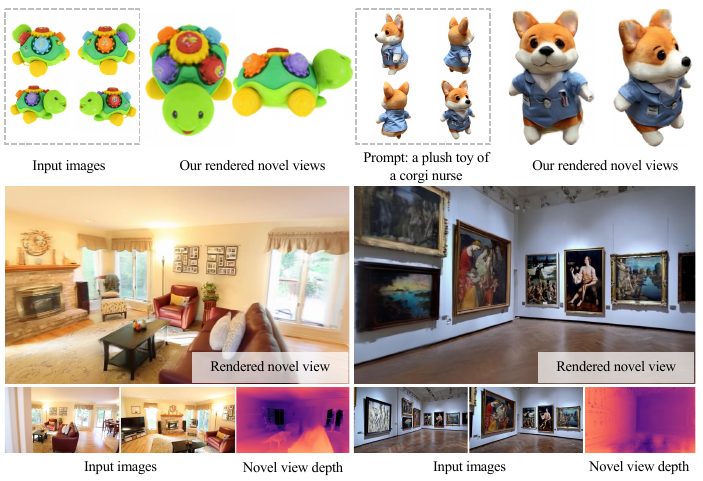}
    \caption{
    Novel-view renderings of our predicted Gaussians from object captures (top left), text-conditioned generated object images (top right), scene captures (bottom left) and text-conditioned generated scene images (bottom right, from \href{https://openai.com/sora}{Sora}~\cite{sora}
    with the prompt ``\textit{Tour of an art gallery with many beautiful works of art in different styles}''
    ). The rendered depth for scenes is at the bottom. By predicting 3D Gaussians with our transformer-based \ourmethod{}, we can naturally handle objects and complex scenes. 
    }
    \label{fig:teaser}
\end{figure}

\let\oldthefootnote\thefootnote
\renewcommand{\thefootnote}{}

\footnotetext{* Equal contribution.}

\let\thefootnote\oldthefootnote

\begin{abstract}
We propose \ourmethod{}, a scalable large reconstruction model that can predict high-quality 3D Gaussian primitives from 2-4 posed sparse images in  $\sim$0.23 seconds on single A100 GPU. Our model features a very simple transformer-based architecture; we patchify input posed images, pass the concatenated multi-view image tokens through a sequence of transformer blocks, and decode final per-pixel Gaussian parameters directly from these tokens for differentiable rendering. In contrast to previous LRMs that can only reconstruct objects, by predicting per-pixel Gaussians, \ourmethod{} naturally handles scenes with large variations in scale and complexity. We show that our model can work on both object and scene captures by training it on Objaverse and RealEstate10K respectively. In both scenarios, the models outperform state-of-the-art baselines by a wide margin. We also demonstrate applications of our model in downstream 3D generation tasks. Our project webpage is available at: \url{https://sai-bi.github.io/project/gs-lrm/}.

\end{abstract}

\input{sections/01_sec_intro}

\input{sections/02_sec_related_work}

\input{sections/03_sec_method}

\input{sections/04_sec_experiments}

\input{sections/05_sec_conclusion}

\vspace{-0.5em}
\section*{Acknowledgement}
We thank Nathan Carr and Duygu Ceylan for useful discussions.

\bibliographystyle{splncs04}
\bibliography{reference}

\input{sections/06_sec_appendix}

\end{document}

%% file: sections/01_sec_intro.tex
\section{Introduction} \label{sec:intro}

Reconstructing a 3D scene from image captures is both a central problem and a long-standing challenge in computer vision.
Traditionally, high-quality 3D reconstruction relies on complex photogrammetry systems \cite{furukawa2009accurate,schonberger2016structure,schoenberger2016mvs} and requires a dense set of multi-view images. Recent advancements in neural representations and differentiable rendering \cite{mildenhall2020nerf,muller2022instant,chen2022tensorf,kerbl20233d} have shown superior reconstruction and rendering quality, by optimizing renderings on a per-scene basis. However, these methods are slow and still require a large number of input views. 
Recently, transformer-based 3D large reconstruction models (LRMs) have been proposed, learning general 3D reconstruction priors from vast collections of 3D objects and achieving sparse-view 3D reconstruction of unprecedented quality \cite{hong2023lrm,li2023instant3d,xu2023dmv3d,wang2023pf}.   
However, these models adopt triplane NeRF \cite{peng2020convolutional,chan2022efficient} as the scene representation, 
which suffers from a limited triplane resolution and expensive volume rendering.  
This leads to challenges in training and rendering speeds, preserving fine details, and scaling to large scenes 
beyond object-centric inputs.

Our goal is to build a general, scalable, and efficient 3D reconstruction model.
To this end, we propose \ourmethod{}, a novel transformer-based large reconstruction model 
that predicts 3D Gaussian primitives~\cite{kerbl20233d} from sparse input images, enabling fast and high-quality rendering and reconstruction for both objects and scenes, as depicted in Fig.~\ref{fig:teaser}.
The core of our approach is a simple and scalable transformer-based network architecture that predicts per-pixel Gaussians. 
Specifically, we patchify input posed images into patch tokens and process them through a series of transformer blocks comprising self-attention and MLP layers, 
and directly regress per-view per-pixel 3D Gaussian primitives from the contextualized multi-view tokens.
Unlike previous LRMs that require careful designs of additional (triplane) NeRF tokens for reconstruction, 
we align input (2D images) and output (3D Gaussians) in the same pixel space, predicting one Gaussian per pixel along the ray. 
This alignment not only simplifies the transformer architecture but also facilitates 3D Gaussians to preserve the high-frequency details in the input images.
Moreover, predicting per-pixel Gaussians allows our model to freely adapt to the input image resolution, 
exhibiting accurate scene details in high-resolution inputs, that previous LRMs with a fixed triplane resolution often struggle with.

Our transformer-based \ourmethod{} is highly scalable from multiple aspects, including model sizes, training data, and scene scales. 
We train two versions of our \ourmethod{} on two large-scale datasets including Objaverse~\cite{objaverse} and Real-Estate10K~\cite{zhou2018stereo}, 
separately for object and scene reconstruction tasks, using the same transformer architecture with minimal domain-specific parameter changes.
The results demonstrate that our \ourmethod{} (with 300M model parameters, up to 16K transformer token length) achieves high-quality sparse-view reconstruction for both 
object and scene scenarios. \textbf{We also achieve state-of-the-art reconstruction quality and outperform previous 
methods by a large margin of 4dB PSNR for objects and 2.2dB PSNR for scenes. }

%% file: sections/02_sec_related_work.tex
\section{Related Work}\label{sec:related_work}

\boldstart{Multi-view 3D Reconstruction.}
3D reconstruction has been extensively studied in computer vision and graphics for decades. 
To address this task, various traditional methods have been proposed, including structure-from-motion (SfM) \cite{pollefeys2004visual,snavely2006photo,agarwal2011building,schonberger2016structure} for sparse reconstruction and calibration, and multi-view stereo (MVS) \cite{pollefeys2008detailed,furukawa2009accurate,schoenberger2016mvs} for dense reconstruction. 
Recently, deep learning-based MVS methods have also been proposed \cite{yao2018mvsnet,yao2019recurrent,gu2020cascade,cheng2020deep,shen2021cfnet}, offering efficient high-quality reconstruction in a feed-forward manner.
Generally, these methods utilize 3D cost volumes — constructed by unprojecting 2D image features into plane sweeps — to achieve high-quality per-view depth estimation.
In this work, we estimate pixel-aligned 3D Gaussian primitives, essentially achieving a per-pixel estimation of depth along with additional Gaussian properties.
Instead of relying on 3D cost volumes, we adopt a multi-view transformer to directly regress Gaussians, using self-attention over all multi-view image patches, naturally allowing for multi-view correspondence reasoning.
Our transformer model can effectively handle highly sparse input views, a big challenge for cost volume-based methods. 

\boldstartspace{Radiance Field Reconstruction.}
Recently, a vast number of works have emerged to address scene reconstruction by optimizing radiance field representations \cite{mildenhall2020nerf} with differentiable rendering, bypassing traditional MVS pipelines.
While NeRF \cite{mildenhall2020nerf} initially represents a radiance field as a coordinate-based MLP, this area has expanded with various models, including voxel-based \cite{liu2020neural,sun2022direct,yu2021pixelnerf}, factorization-based \cite{chen2022tensorf,chan2022efficient,chen2023factor,chen2023dictionary}, hash grid-based\cite{muller2022instant}, and point- (or primitive-) based \cite{xu2022point,lombardi2021mixture,gao2023strivec,kerbl20233d} representations among others \cite{barron2022mip,barron2021mip,barron2023zip,zhang2020nerf++}.
We leverage Gaussian Splatting \cite{kerbl20233d}, a state-of-the-art technique for radiance field modeling and rendering, allowing us to achieve real-time rendering and large-scale scene reconstruction.
In contrast to these radiance field models optimized on a per-scene basis given dense images, we propose a generalizable model trained across scenes for feed-forward sparse-view reconstruction.

\boldstartspace{Feed-forward Reconstruction.} 
Prior works have proposed various feed-for-ward 3D reconstruction and rendering methods.
Early efforts adopt CNNs to estimate neural points \cite{aliev2020neural,wiles2020synsin,Yifan2019DSS} or multi-plane images (MPIs) \cite{zhou2018stereo,mildenhall2019local,lin2020deep}, achieving rendering via point splatting or alpha compositing. 
We utilize Gaussian Splatting, which can be seen as a generalization of point splatting and MPI compositing.
Recently, generalizable radiance field-based methods have been proposed and achieved state-of-the-art quality \cite{yu2021pixelnerf,wang2021ibrnet,suhail2022light,chen2021mvsnerf}, leveraging NeRF-like volume rendering. 
These methods typically leverage 3D-to-2D geometric projection to sample per-view 2D features for 3D reconstruction, using model designs like epipolar lines \cite{wang2021ibrnet,suhail2022light,suhail2022generalizable} or plane-swept cost-volumes \cite{chen2021mvsnerf,johari2022geonerf,long2022sparseneus} similar to MVS. 
We instead leverage a clean large transformer model without such 3D inductive bias, and directly regress per-view pixel-aligned 3D Gaussian primitives.
The dense self-attention layers in our model can effectively learn multi-view correspondence and general reconstruction priors, leading to significantly better rendering quality than previous epipolar-based methods (see Tab.~\ref{tab:compare_scenes}).

Our model is inspired by the recent transformer-based 3D large reconstruction models (LRMs) \cite{hong2023lrm,li2023instant3d,xu2023dmv3d,wang2023pf}, which are based on triplane NeRF and focus on objects.
We propose a novel LRM for Gaussian Splatting with a simplified network architecture and per-pixel Gaussian prediction mechanism, achieving better quality, faster rendering, and scaling up to handle large scenes.

PixelSplat \cite{charatan2023pixelsplat} and LGM \cite{tang2024lgm} are two concurrent works that are also based on pixel-aligned 3D Gaussian prediction. In particular, LGM leverages a U-Net architecture and only focuses on object generation; PixelSplat leverages epipolar line-based sampling and only tackles scene-level reconstruction. In contrast, our \ourmethod{} is a clean transformer model that is much simpler to implement and scale. We demonstrate that our model significantly outperform these two concurrent works in terms of both object and scene-level 
reconstructions.

%% file: sections/03_sec_method.tex
\section{Method}\label{sec:method}

In this section, we present the technical details of our method, including the architecture of 
our transformer-based model (\secref{sec:arch}) and the loss functions (\secref{sec:loss}).

\begin{figure}[t]
    \centering
    \includegraphics[width=0.99\textwidth]{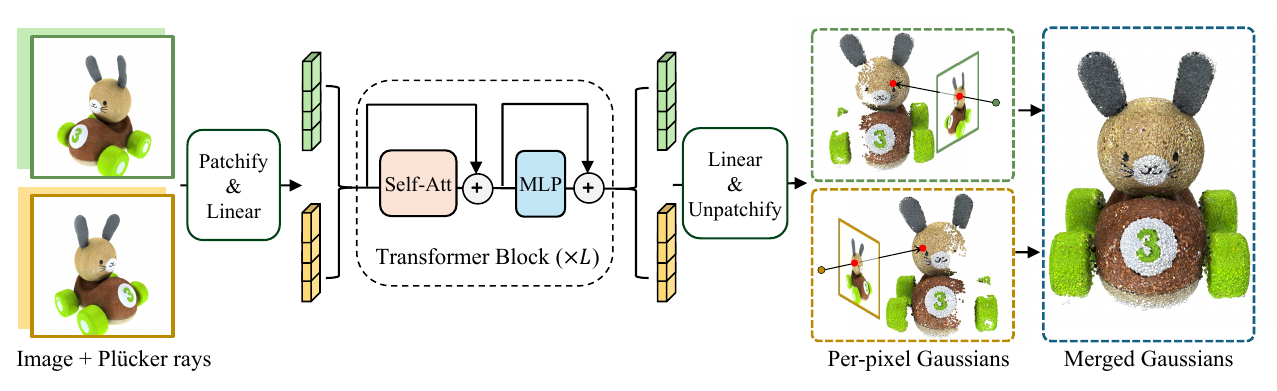}
    \caption{
    Our simple transformer-based \ourmethod{} predicts 3D Gaussian parameters from sparse posed images. 
    Images are patchified and the concatenated patch tokens are sent to the transformer blocks.
    By unpatchifying the transformer's output,
    each pixel is unprojected to a 3D Gaussian.
    The final output merges all 3D Gaussians. (Note that here we visualize the Gaussian centers and colors as point clouds for illustration; please refer to in Fig.~\ref{fig:teaser} for the splatting-based renderings.)
    }
    \label{fig:pipeline}
\end{figure}

\subsection{Transformer-based Model Architecture}\label{sec:arch}
As shown in \figref{fig:pipeline}, we train a transformer model to regress per-pixel 3D GS parameters from 
a set of images with known camera poses. 
We tokenize posed input images via a patchify operator~\cite{dosovitskiy2020image}.
Multi-view image tokens are then concatenated and passed through a sequence of transformer blocks consisting of self-attention and MLP layers.
From each output token, we decode the attributes of pixel-aligned Gaussians in the corresponding patch with a linear layer. 

\boldstartspace{Tokenizing posed images.}
The inputs to our model are $N$ multi-view images $\{\img_i \in \mathbb{R}^ {H\times W \times 3}  | i = 1,2,..,N\}$
and their camera intrinsic and extrinsic parameters; here $H$ and $W$ are the height and width of the images.
Following prior works~\cite{xu2023dmv3d, chen2023:ray-conditioning}, 
we use the Pl\"ucker ray coordinates~\cite{plucker1865xvii} of each image $\{\mathbf{P}_i \in \mathbb{R}^{H \times W \times 6} \}$ computed from the camera parameters for pose conditioning. 
Specifically, we concatenate the image RGBs and their Pl\"ucker coordinates channel-wise, enabling per-pixel pose conditioning and forming a per-view feature map with 9 channels.
Similar to ViT~\cite{dosovitskiy2020image},  
we patchify the inputs by dividing the per-view feature map into non-overlapping patches with a patch size of $p$. For each 2D patch, we flatten it 
into a 1D vector with a length of $p^2 \cdot 9$. Finally, we adopt a linear layer that
maps the 1D vectors to image patch tokens of $d$ dimensions, where $d$ is the transformer width. 
Formally, this process can be written as:
\begin{align}
\{\tok_{ij}\}_{j=1,2,...,HW/p^2}=\text{Linear}\big(\text{Patchify}_{p}\big(\text{Concat}(\img_i, \mathbf{P}_i)\big)\big),
\end{align}
where $\{\tok_{ij} \in \mathbb{R}^{d}\}$ denotes the set of patch tokens  for image $i$, and 
there are a total number of $H  W/p^2$  such tokens (indexed by $j$) for each image. 
As Pl\"ucker coordinates vary across pixels and views, they naturally serve as spatial embeddings to distinguish different patches; hence we do not use 
additional positional embedding as in~\cite{dosovitskiy2020image} or view embeddings~\cite{wang2023pf}.

\boldstartspace{Processing image tokens with transformer. } 
Given the set of multi-view image tokens $\{\tok_{ij}\}$, we concatenate and feed them through a chain of 
transformer blocks~\cite{vaswani2017attention}:  
\begin{align}
    &\{\tok_{ij}\}^{0} = \{\boldsymbol{T}_{ij}\}, \\
    &\{\tok_{ij}\}^{l} = \text{TransformerBlock}^{l}(\{\boldsymbol{T}_{ij}\}^{l-1}), l=1,2,..., L,
\end{align}
where $L$ is the total number of transformer blocks. 
Each  transformer block 
is equipped with residual connections~\cite{he2016deep} (i.e., the operator `+' in Fig.~\ref{fig:pipeline}) and consists of Pre-LayerNorm~\cite{ba2016layer}, 
multi-head Self-Attention~\cite{vaswani2017attention} and MLP. 

\boldstartspace{Decoding output tokens to per-pixel Gaussians.} 
With the output tokens $\{\tok_{ij}\}^L$ from the transformer, we decode them into  Gaussian parameters using a single linear layer:
\begin{align}
    \{\mathbf{G}_{ij} \} = \mathrm{Linear} (\{\tok_{ij}\}^L),  
\end{align}
where $\mathbf{G}_{ij} \in \mathbb{R}^{p^2 \cdot q}$ represents the 3D Gaussian and $q$ is the number of parameters per Gaussian.
We then unpatchify $\mathbf{G}_{ij}$ into $p^2$ Gaussians.
Since we use the same patch size $p$ for patchifying and unpatchifying operations, 
we end up with $HW$ Gaussians for each view where each 2D pixel corresponds to one 3D Gaussian.

Similar to \cite{kerbl20233d}, the 3D Gaussian is parameterized by 3-channel RGB, 3-channel scale, 4-channel rotation quaternion, 1-channel opacity, and 1-channel ray distance. 
Thus $q\mbox{=}12$ in our formulation.
For splatting rendering, the location of the Gaussian center is obtained by the ray distance and the known camera parameters.
Specifically, suppose that $t$, $\mathrm{ray}_o$, $\mathrm{ray}_d$ are the ray distance, ray origin, and ray direction, respectively, the Gaussian center 
$\mathrm{xyz} = \mathrm{ray}_o + t \cdot  \mathrm{ray}_d$.
We will discuss the detailed initialization of these parameters in the Appendix.

The final output of our model is simply the merge of 3D Gaussians from all $N$ input views. Thus the model will output $N \cdot HW$ Gaussians in total. It's worth noting that the number of Gaussians scale up with increased input resolution, which is in contrast to the fixed-resolution triplane used in prior LRM works~\cite{hong2023lrm, li2023instant3d, xu2023dmv3d, wang2023pf}. This property allows us to better handle high-frequency details in the inputs and large-scale scene captures.

\subsection{Loss Functions}\label{sec:loss}
During training, we render the images at the $M$ supervision views using the predicted Gaussian splats, and minimize the image reconstruction loss.
Let $\{\img_{i'}^* | i' = 1, 2,..., M\}$ be the set of groundtruth views, and $\{\hat{\img}_{i'}^*\}$ be the rendered images, our loss function 
is a combination of MSE (Mean Squared Error) loss and Perceptual loss:
\begin{align}
    \mathcal{L} = \frac{1}{M} \sum_{i'=1}^M\left(\mathrm{MSE}\left(\hat{\img}^*_{i'},\img^*_{i'}\right) 
    +\lambda \cdot \mathrm{Perceptual}\left(\hat{\img}^*_{i'},\img^*_{i'}\right) \right),
\end{align}
where $\lambda$ is the weight of the perceptual loss. We empirically find that the perceptual loss   in~\cite{chen2017photographic}
based on VGG-19 network~\cite{simonyan2014very} 
provides a more stable training than LPIPS~\cite{zhang2018perceptual} used in  \cite{hong2023lrm,li2023instant3d, xu2023dmv3d, wang2023pf}, and we use it in this work.

%% file: sections/04_sec_experiments.tex
\section{Experiments} \label{sec:experiments}

\input{tables/comp_scene}

In this section, we first describe the training and testing datasets (\secref{sec:dataset}), 
then introduce the implementation and training details (\secref{sec:impl}). We make both quantitative 
and qualitative comparisons (\secref{sec:comp})  against different baselines. 
Finally we show some downstream applications (\secref{sec:app}) of our method. 
\textbf{We refer the readers to \href{https://sai-bi.github.io/project/gs-lrm/}{our project page} for video results and interactive visualizations.}

\subsection{Datasets}~\label{sec:dataset} 

\boldstart{Object-level dataset.} 
We use the Objaverse dataset~\cite{objaverse} to train our object-level reconstruction model. 
We only leverage the multi-view renderings of the objects without accessing explicit 3D information (such as depths).
Following~\cite{hong2023lrm}, we center and scale each 3D object to 
a bounding box of $[-1, 1]^3$, and render 32 views randomly placed 
around the object with a random distance in the range of $[1.5, 2.8]$.
Each image is rendered at a resolution of $512 \times 512$ under uniform lighting.
We render a total of $730$K objects. 
We evaluate our model on two 3D object
datasets including the full Google Scanned Objects (GSO)~\cite{gso} that contains 1009 objects and 
the Amazon Berkeley Objects (ABO)~\cite{abo} dataset from which we sample 1000 objects. 
We follow Instant3D~\cite{li2023instant3d} and render 4 structured input views 
evenly placed around the objects with an elevation of $20^{\circ}$ to ensure a good coverage,
and randomly select another $10$ views for testing.

\boldstartspace{Scene-level dataset.} 
We use the RealEstate10K~\cite{zhou2018stereo} dataset to train our scene-level model. 
It sources from real estate video footages and has both indoor and outdoor scenes.
The dataset contains $80$K video clips curated from $10$K YouTube videos.
Each clip has a set of extracted frames with known camera poses estimated by SfM~\cite{schoenberger2016sfm}. 
We follow the standard training/testing split for the dataset, which is also used in pixelSplat~\cite{charatan2023pixelsplat}.

\subsection{Implementation  Details}\label{sec:impl}
We have two models trained independently in this paper: object-level \ourmethod{} and scene-level \ourmethod{}.
The two models share the same model architecture and take almost the same training recipe.
The differences are in the training data (Sec.~\ref{sec:dataset}) and view selection and normalization (see details below). We also made necessary changes for fair comparisons with baseline methods (Sec.~\ref{sec:comp}). 

\boldstartspace{Model details.} 
We use a patch size of $8 \times 8$ for the image tokenizer. Our transformer consists of 
24 layers, and the hidden dimension of the transformer is $1024$. Each transformer block 
consists of a multi-head self-attention layer with $16$ heads, 
and a two-layered MLP with GeLU activation. 
The hidden dimension of the MLP is $4096$.
Both layers are equipped with Pre-Layer Normalization (LN) and residual connections.
Besides the above Pre-LNs, Layer Normalization is used after the patchifying Linear layer and before the unpatchifying Linear layer to stabilize the training. 

\boldstartspace{Training details.} To enable efficient training and inference, we adopt Flash-Attention-v2~\cite{dao2023flashattention} in the xFormers~\cite{xFormers2022} library, gradient checkpointing~\cite{chen2016training}, and mixed-precision training~\cite{micikevicius2018mixed} with BF16 data type. 
We also apply deferred backpropagation~\cite{zhang2022arf} for 
rendering the GS to save GPU memory.
We pre-train the model with a resolution of $256\times 256$ and fine-tune the trained model with a resolution of 
$512 \times 512$ for a few epochs.  
The fine-tuning shares the same model architecture and initializes the model with the pre-trained weights, but processes more tokens than the pre-training.
At each training step, for object-level, we sample a set of $8$ images (from $32$ renderings) as a data point, and from which we randomly select $4$ input views and $4$ supervision views independently. 
This sampling strategy encourages more overlap between input views and rendering views than directly sampling from $32$ rendering views, which helps the model's convergence.
For scene-level, we adopt two input views for a fair comparison with pixelSplat \cite{charatan2023pixelsplat}. 
Following pixelSplat \cite{charatan2023pixelsplat}, we select $2$ random input views and then randomly sample supervision views between them; we use $6$ supervision views for each batch. 
We normalize the camera poses for scene-level input images following common practices in 
previous forward-facing reconstructions as done in~\cite{mildenhall2019llff,chen2022tensorf}. 
We further fine-tune a model that takes $2-4$ input images of $512\times 512$ for generating visual results.  
For both models, we use 64 A100 (40G VRAM) GPUs to train our models; 256-res pretraining takes 2 days, while 512-res finetuning takes 1 additional day.
For more details, please refer to the Appendix.

\begin{figure}[t]
    \centering
    \includegraphics[width=0.98\textwidth]{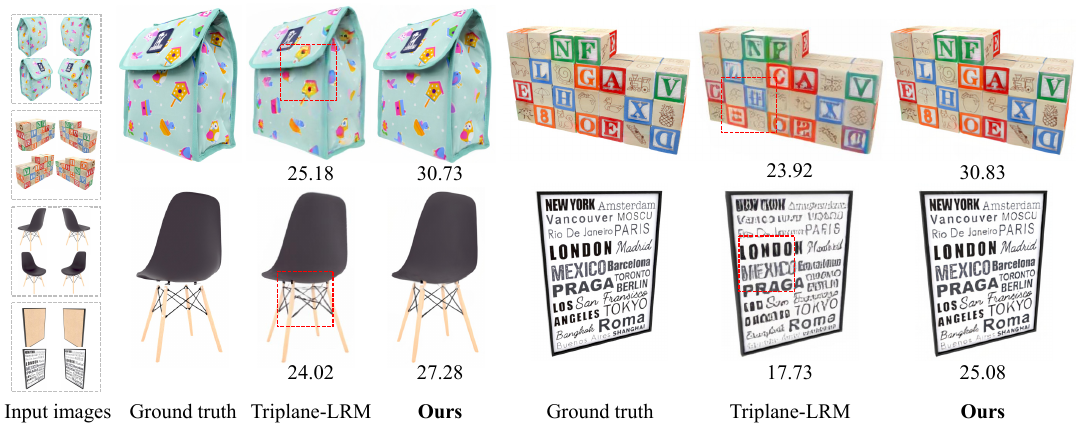}
    \caption{Visual comparisons to Instant3D's Triplane-LRM~\cite{li2023instant3d}. 
    The 4-view input images are shown in the leftmost column, and we compare novel view renderings on the right.
    The Triplane-LRM cannot reconstruct high-frequency details (top left and top right) and thin structures (bottom left) well. It also suffers from texture distortion artifacts (bottom right), possibly due to a lack of representation capability and the non-pixel-aligned prediction of triplanes. In contrast, our \ourmethod{} works significantly better in these cases.
    PSNRs are shown under each image.
    }  
    \label{fig:obj_results_compare_instant3d}
\end{figure}

\begin{figure}[ht!]
    \centering
    \includegraphics[width=0.98\textwidth]{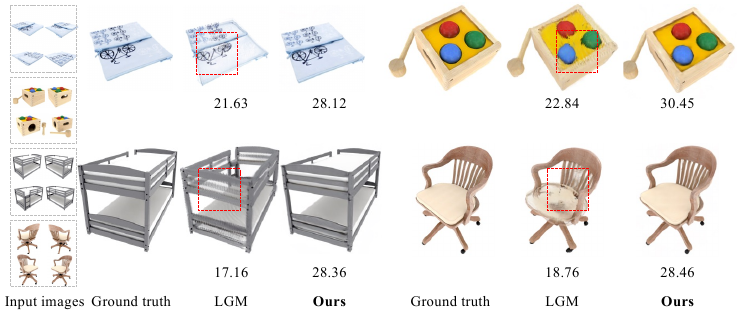}
    \caption{
    Visual comparisons to LGM~\cite{tang2024lgm}.
    The LGM renderings have obvious distorted textures (top) and broken geometries (bottom) and are inferior in recovering accurate surface opacity (top left; bottom left; bottom right). 
    Our \ourmethod{} renderings recover the high-frequency details and are visually much closer to the ground truth.
    PSNRs are shown under each image. 
    } 
    \label{fig:obj_results_compare_lgm}
\end{figure}

\subsection{Evaluation against Baselines}~\label{sec:comp}

\noindent\textbf{Object-level.} We compare our object-level \ourmethod{} with the Triplane-LRM in Instant3D~\cite{li2023instant3d}. We outperform this baseline by a large margin across all view synthesis metrics; for example, as shown in Tab.~\ref{tab:compare_objects}, we improve the PSNR for novel-view rendering by 3.98dB on GSO data, and by 1.59dB on ABO data.  This is also reflected by our much sharper renderings in Fig.~\ref{fig:obj_results_compare_instant3d}; our \ourmethod{} manages to faithfully reproduce the high-frequency details, e.g., texts, in the input images, while Triplane-LRM tends to blur out the details. 
We attribute this to our pixel-aligned Gaussian prediction scheme which creates a shortcut for learning accurate per-Gaussian colors from input RGB images; this is in contrast to the non-pixel-aligned prediction of triplanes in Instant3D's LRM where the relationship between input pixel colors and triplane features is less straightforward, and might be more challenging to learn for the network. 
Another advantage of our \ourmethod{} is that our predicted Gaussians are much faster to render than the predicted NeRF from Triplane-LRM, making it easier to deploy in downstream applications.
We also tried to compare against another baseline SparseNeuS~\cite{long2022sparseneus}; however, we found that it failed to produce plausible reconstructions given 4 highly sparse inputs; this was also observed in the prior Instant3D work (they had to use 8 views to run SparseNeuS, which is not a fair comparison).

In addition, we compare with the concurrent work LGM~\cite{tang2024lgm} which is also based on Gaussian Splatting~\cite{kerbl20233d} . 
The official LGM is trained with a special setting using $256\!\times\!256$ resolution input and $512\!\times\!512$ resolution output supervision.
Since their model only accepts $256\!\times\!256$ input, we compare with LGM using our low-res model, trained with $256\!\times\!256$ images only from our 256-res pre-training stage.
We evaluate both models with $256\!\times\!256$ renderings for comparison.
As seen in Tab.~\ref{tab:compare_objects}, our approach significantly outperforms LGM, achieving a notable 8dB higher PSNR on both GSO and ABO testing data. 
The improvement can be visualized in Fig.~\ref{fig:obj_results_compare_lgm}.
It's worth noting that this is an almost equal-compute comparison: LGM is trained on 32 A100 (80G VRAM) for 4 days, while our low-res base model is trained using 64 A100 (40G VRAM) for 2 days. 
This further highlights the method-wise advantage of our \ourmethod{} --- a transformer model predicting per-pixel Gaussians that scales up easily with data and compute.

\begin{figure}[t!]
    \centering
    \includegraphics[width=0.98\textwidth]{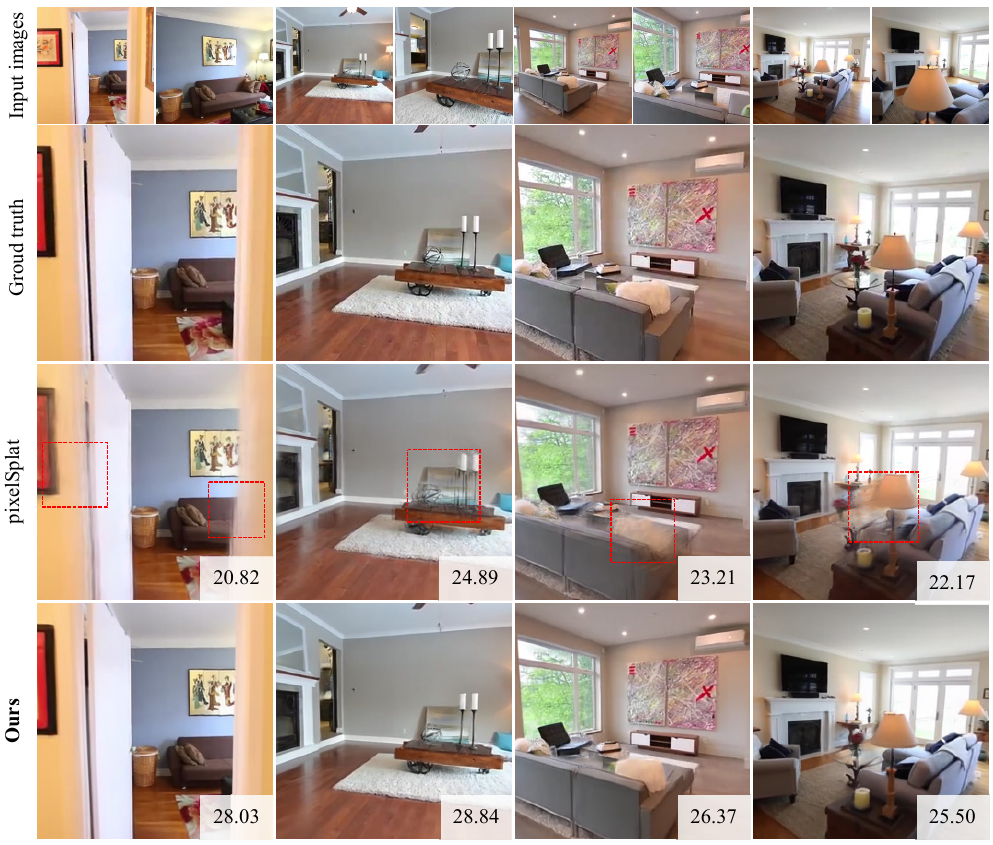}
    \caption{We compare scene-level \ourmethod{} with the best-performing baseline pixelSplat~\cite{charatan2023pixelsplat}.
    We can observe that our model is better in sharpness (leftmost column), has fewer floaters (mid-right and rightmost), and is more faithful to the original scenes (mid-left).
    Our superiority in visual quality aligns with the significant quantitative metric improvement in Tab.~\ref{tab:compare_scenes}.
    PSNRs are shown at the corner of each image. 
    } 
    \label{fig:scene_results}
\end{figure}

\medskip\noindent\textbf{Scene-level.} We compare our scene-level \ourmethod{} against previous generalizable neural rendering techniques~\cite{yu2021pixelnerf,suhail2022generalizable,du2023learning} and the 
state-of-the-art GS-based concurrent work pixelSplat~\cite{charatan2023pixelsplat}. Since pixelSplat and GPNR are trained at $256\times 256$ image resolution, we use our low-res model in these comparisons, and follow exactly the same evaluation setting as in pixelSplat. We directly take the reported quantitative results for all baselines from pixelSplat~\cite{charatan2023pixelsplat}. 
As shown in Tab.~\ref{tab:compare_scenes}, our approach achieves the best quantitative results on the RealEstate10k testing set, substantially surpassing all baselines for all metrics. In particular, when compared to pixelSplat, the top-performing baseline, our model leads to significant improvements of 2.2db in PSNR, 0.034 in SSIM, and 0.028 in LPIPS.
These metric improvements align with the visual comparisons in Fig.~\ref{fig:scene_results}, where our results are sharper and have fewer floaters.
Note that pixelSplat and other baseline methods all leverage more complex model designs such as epipolar line-based feature aggregation.
In contrast, we utilize a straightforward transformer model with self-attention layers.
These self-attention layers effectively learn to aggregate context information from all the intra-view and inter-view pixels (as opposed to a subset of them on epipolar lines) for accurate per-pixel Gaussian prediction, when trained on large amount of data.

\subsection{High-resolution Qualitative Results}\label{sec:more_results}
\begin{figure}[ht!]
    \centering
    \includegraphics[width=0.98\textwidth]{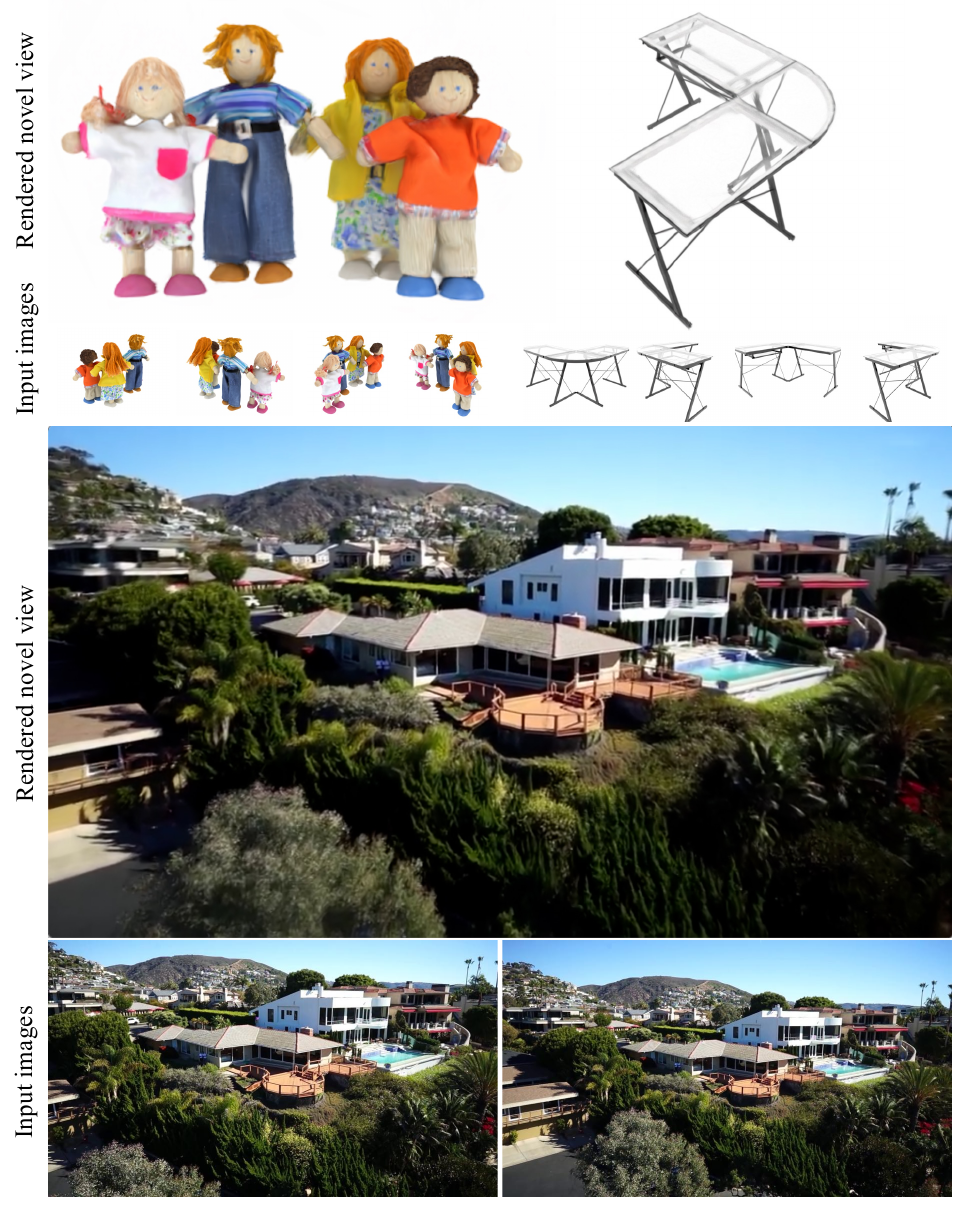}
    \vspace{-0.5em}
    \caption{
    We show high-res novel-view renderings  from our predicted GS given high-res input images (4 $512\!\times\!512$ images for objects, and 2 $512\!\times\!904$ images for a scene; rendering resolution is the same to the input), 
    to demonstrate our \ourmethod{}'s capability to represent fine-grained details, e.g., readable texts (top left), 
    translucent and thin structures (top right, bottom). 
    Image sources are from GSO (top left), ABO (top right), and RealEstate10K (bottom). 
    }
    \label{fig:highres_obj}
\end{figure}

We showcase some high-res reconstructions of our \ourmethod{} in Fig.~\ref{fig:highres_obj}. For the object captures, texts on the top-left product box remain readable in our rendered novel views, even when the inputs are captured from oblique viewpoints; we also manage to reconstruct the challenging thin structures and transparent glasses in the top-right table example. For the scene captures, we are able to handle large outdoor depth variations and faithfully reproduce complex structures, e.g., trees, in the presented examples. 
Please refer to \href{https://sai-bi.github.io/project/gs-lrm/}{our project page} for videos and interactive rendering results.

\subsection{Applications in 3D Generation}\label{sec:app}
Following the Instant3D~\cite{li2023instant3d} work, we can also chain our \ourmethod{} with a text-conditioned or image-conditioned multi-view generator to achieve text-to-3D or image-to-3D. We qualitatively show some results to demonstrate such a workflow for applying our models in this downstream 3D generation task.  

\begin{figure}[t]
    \centering
    \includegraphics[width=0.98\textwidth]{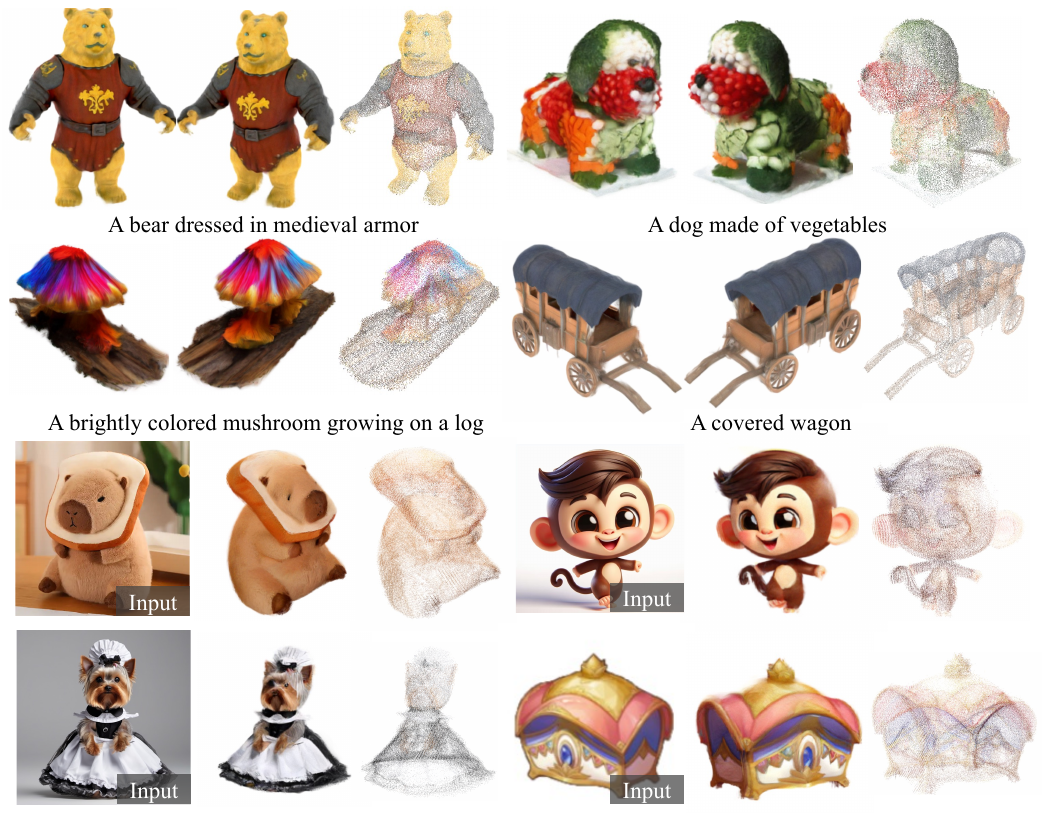}
    \caption{Text-to-3D (top two rows) and image-to-3D (bottom two rows) results by chaining Instant3D's~\cite{li2023instant3d} text-conditioned and Zero123++'s~\cite{shi2023zero123plus} image-conditioned multi-view generators to our \ourmethod{} reconstructor. 
    For each result, we show novel view renderings and a visualization of the point cloud with point positions and colors extracted from the predicted Gaussians. 
    }
    \label{fig:text-to-3D}
\end{figure}

\medskip\noindent\textbf{Text/Image-to-object.} For the text-to-3D application, we use the finetuned SDXL~\cite{podell2023sdxl} model in Instant3D as the text-to-multi-views generator. 
Since the above generator generates four structured views with known camera parameters, we directly feed these posed images into our object-level \ourmethod{} to get 3D GS instantly. 
The results are visualized in Fig.~\ref{fig:text-to-3D} (top two rows).
We provide both novel view renderings and point clouds for illustrating the appearance and geometry, respectively. For the image-to-3D application, we use the image-conditioned multi-view diffusion model in Zero123++~\cite{shi2023zero123plus}, which generates 6 structured views at fixed viewpoints. Though being trained with 4 input views, our transformer-based \ourmethod{} can take a variable number of images, e.g., 6 images, during inference; hence we simply input the 6 generated images (along with their cameras) into our object-level \ourmethod{} to predict the 3D Gaussians. We also show our novel-view renderings and a point cloud visualization of 3D Gaussians in Fig.~\ref{fig:text-to-3D} (bottom two rows).

\begin{figure}[t]
    \centering
    \includegraphics[width=0.98\textwidth]{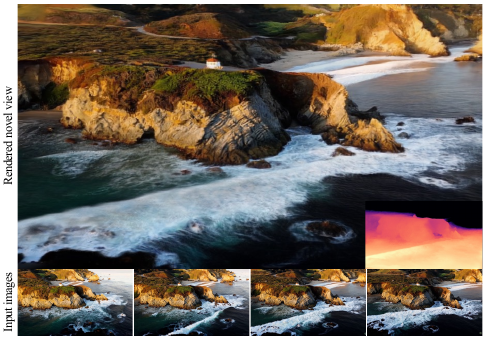}
    \caption{Text-to-scene result. We take the generated video from the Sora text-to-video model~\cite{sora}
    (Prompt: \textit{drone view of waves crashing against the rugged cliffs along Big Sur’s garay point beach}).
    We select 4 frames and reconstruct the 3D scene from them using our scene-level \ourmethod{}. 
    We show novel-view RGB/depth rendering from our predicted GS (top) and 4 input images (bottom).
    Please refer to \href{https://sai-bi.github.io/project/gs-lrm/}{our project page} for the video and interactive rendering results.
    } 
    \label{fig:text-to-scene}
\end{figure}

\medskip\noindent\textbf{Text-to-scene.} 
 We adopt the recently proposed Sora video generation model~\cite{sora} as the multi-view scene generator from texts. 
As the Sora model has not publicly released, 
we use the generated videos published in~\cite{sora}. 
Our current model is limited to static scenes only, and we thus pick the generated videos from relevant text prompt guidance.
We use COLMAP~\cite{schoenberger2016sfm} to register the video frames, then feed a selected sparse subset of posed frames into our scene-level \ourmethod{} to reconstruct the 3D scene. 
The visualization is in Fig.~\ref{fig:text-to-scene}.

\subsection{Limitations}
Although our method shows high-quality reconstruction results from posed sparse images, there are still a few limitations to be addressed in future work. 
Firstly, the highest resolution our model can currently operate on is about $512 \times 904$; it is of significant interest to extend the model to work on 1K, even 2K resolution images for best visual quality. 
Secondly, our model requires known camera parameters; this assumption may not be practical in certain application scenarios (e.g., a user only captures 4 views around an object, making it extremely challenging for SfM to work). To make our model more practical, it is interesting to explore ways to get rid of the camera parameters from the input end~\cite{wang2023pf,wang2023dust3r,jiang2022LEAP}. 
Thirdly, the pixel-aligned representation only explicitly models the surface inside the view frustum, which means that unseen regions cannot be reconstructed.
We found that the model has a certain ability to re-purposing points for hallucinating unseen parts (also observed in \cite{szymanowicz2023splatter_image}), but this capacity is limited and not guaranteed.
We leave it to future work to improve the unseen regions.

%% file: tables/comp_scene.tex
\begin{table}[t]
\caption{Comparison against baselines on object-level (left) and scene-level (right) reconstructions. We matched the baseline settings by comparing with Instant3D's Triplane-LRM~\cite{li2023instant3d} and LGM~\cite{tang2024lgm} at $512$ and $256$ resolution respectively for both input and rendering. All scene-level results were performed at $256$ resolution for fair comparisons. We outperform relevant baselines by a large margin in both scenarios.
}
\begin{minipage}{.6\textwidth}

\resizebox{0.99\linewidth}{!}{
\begin{tabular}{r|ccc|ccc } 
    & \multicolumn{3}{c|}{GSO} &  \multicolumn{3}{c}{ABO}  \\ 
    & PSNR \(\uparrow\) & SSIM \(\uparrow\)  & LPIPS \(\downarrow\) & PSNR \(\uparrow\) & SSIM \(\uparrow\) & LPIPS \(\downarrow\) \\ 
    \midrule
    Triplane-LRM~\cite{li2023instant3d} &  26.54          & 0.893         & 0.064     & 27.50 & 0.896  & 0.093  \\
    \textbf{Ours} (Res-512) & \textbf{30.52} & \textbf{0.952} & \textbf{0.050} &  \textbf{29.09}  &  \textbf{0.925} & \textbf{0.085}   \\
    \midrule
    LGM~\cite{tang2024lgm} & 21.44          & 0.832         & 0.122    & 20.79 & 0.813  & 0.158 \\
    \textbf{Ours} (Res-256)  & \textbf{29.59} & \textbf{0.944} & \textbf{0.051} &  \textbf{28.98}  &  \textbf{0.926} & \textbf{0.074}   \\
\end{tabular}
}
\vspace{0.02cm}
\label{tab:compare_objects}

\end{minipage}%
\begin{minipage}{.42\textwidth}
\resizebox{0.99\linewidth}{!}{
\begin{tabular}{r|ccc} 
    &  \multicolumn{3}{c}{RealEstate10k}  \\ 
    & PSNR \(\uparrow\) & SSIM \(\uparrow\)  & LPIPS \(\downarrow\)  \\ 
    \midrule
    pixelNeRF~\cite{yu2021pixelnerf} & 20.43 & 0.589  & 0.550 \\
    GPNR~\cite{suhail2022generalizable} & 24.11 & 0.793 &  0.255 \\
    Du et. al~\cite{du2023learning} & 24.78 & 0.820  & 0.213 \\
    pixelSplat~\cite{charatan2023pixelsplat} & {25.89} & {0.858}  & {0.142} \\
    \textbf{Ours}  &  \textbf{28.10}  &  \textbf{0.892} & \textbf{0.114}  \\
\end{tabular}
}
\vspace{0.06cm}

\label{tab:compare_scenes}

\end{minipage}%

\end{table}

%% file: sections/05_sec_conclusion.tex
\section{Conclusion}\label{sec:conclusion}
In this work, we present a simple and scalable transformer-based large reconstruction model for Gaussian Splatting (GS) representation. Our method enables fast feed-forward high-res GS prediction from a sparse set of posed images in $\sim$0.23 seconds on a single A100 GPU. Our model can work on both object-level and scene-level captures, and achieves state-of-the-art performance in both scenarios when trained on large amount of data. We hope that our work can inspire more future work in the space of data-driven feed-forward 3D reconstruction.

%% file: sections/06_sec_appendix.tex
\newpage

\appendix
\section{Appendix}

\subsection{Pseudo Code}
We list the pseudo code of our \ourmethod{} in Algorithm~\ref{alg:code}~\footnote{To avoid the the shaping constraints of the CUDNN kernel, we use Einops~\cite{rogozhnikov2021einops} and Linear layer to implement the patchify and unpatchify operator, but they are conceptually the same to the conv/deconv operator. 
For clarity, we use conv/deconv in pseudo code.}.
The code implements the method that we discussed in the main method section, and also the Gaussian parametrization detailed later  in Sec.~\ref{sec:appen_gaussian_parameterization}.

\subsection{Additional Model Details}
We do not use bias term throughout our model, which includes both Linear and LayerNorm layers. 
We initialize the model weights with a normal distribution of zero-mean and $0.02$-stddev.

\subsection{Additional Training Details}
We train our model with AdamW~\cite{loshchilov2017decoupled} optimizer.
The $\beta_1$, $\beta_2$  are set to $0.9$ and $0.95$ respectively.
We use a weight decay of $0.05$ on all parameters except the those of the LayNorm layers.
We use a cosine learning rate decay with linear warmup.
We take $2000$ step of warm up and the peak learning rate is set to $4e-4$.
The model is trained for $80$K iterations on the $256$-res training, and then fine-tuned with $512$-res for another $20$K iterations. Finetuing at $512$-res uses $500$-step warmup with the peak learning rate $1e-4$ in the cosine learning rate decay schedule. We use a per-GPU batch size of 8 objects/scenes during 256-res training, and a per-GPU batch size of 2 during 512-res finetuning. For each object, we use 4 input views and 4 novel supervision views at each iteration of 256-res and 512-res training; for each scene, we use 2 input views and 6 novel supervison views, following the protocol of pixelSplat.
We use $\lambda=0.5$ to balance the MSE loss and Perceptual loss.
To enable efficient training and inference, we adopt Flash-Attention-v2~\cite{dao2023flashattention} in the xFormers~\cite{xFormers2022} library, gradient checkpointing~\cite{chen2016training}, and mixed-precision training~\cite{micikevicius2018mixed} with BF16 data type. 
We also apply deferred backpropagation~\cite{zhang2022arf} for rendering the GS to save GPU memory. 256-res training takes about 2 days on 64 A100 (40G VRAM) GPUs, while 512-res finetuning costs 1 additional day.

\begin{algorithm}[t]

\caption{GS-LRM  pseudo code.}\label{alg:code}
\newcommand{\hlbox}[1]{%
  \fboxsep=1.2pt\hspace*{-\fboxsep}\colorbox{blue!10}{\detokenize{#1}}%
}
\lstset{style=mocov3}
\vspace{-3pt}
\begin{lstlisting}[
    language=python,
    escapechar=@,
    label=code:gslrm]
    # Input list: 
    # image: [b, n, h, w, 3], n is number of views; h and w are the height and width
    # extrinsic: [b, n, 4, 4] 
    # intrinsic: [b, n, 4] 
    # 
    # Output list (3D GS parameters):
    # xyz: [b, *, 3], rgb [b, *, 3], scaling [b, *, 3], rotation [b, *, 4], opacity [b, *, 1]

    # GS-LRM transformer
    o, d = rays_from_camera(extrinsic, intrinsic)   # rays_origin, rays_direction: [b, n, h, w, 3]
    x = concat([image, o, cross(o, d)], dim=-1)     # [b, n, h, w, 9]
    x = conv(x, out=d, kernel=8, stride=8)          # patchify to [b, n, h/8, w/8, d]
    x = x.reshape(b, -1, d) # Sequentialize as transformer input [b, n * h/8 * w/8, d]
    x = transformer(LN(x))                          
    x = x.reshape(b, n, h//8, w//8, d)
    x = deconv(LN(x), out=12, kernel=8, stride=8)    # unpatchify to GS output: [b, n, h, w, 12]
    x = x.reshape(b, -1, 12)    # Simply merge all Gaussians together [b, n * h * w, 12]

    # GS parameterization
    distance, rgb, scaling, rotation, opacity = x.split([1, 3, 3, 4, 1], dim=-1)
    w = sigmoid(distance)
    xyz = o + d * (near * (1 - w) + far * w)
    scaling = min(exp(scaling - 2.3), 0.3)
    rotation = rotation / rotation.norm(dim=-1, keepdim=True)
    opacity = sigmoid(opacity - 2.0)

    return xyz, rgb, scaling, rotation, opacity

\end{lstlisting}\vspace{-5pt}
\end{algorithm}

\subsection{3D Gaussian Parameterization}
\label{sec:appen_gaussian_parameterization}
As 3D Gaussians are unstructured explicit representation (i.e., different from Triplane-NeRF's structural implicit representation), the parameterization of the output parameters can largely affect the model's convergence.
For here, the `structural' and `unstructual' mainly refer to whether each token have a determinitic spatial meaning.
We discuss in detail how we implement the parameterization for reproducibility.

\boldstartspace{Scale and opacity.} For the scale and opacity of the Gaussian Splatting, we apply the default activations used by Gaussian Splatting~\cite{kerbl20233d} to map the range to $\mathbb{R}^{+}$ and $(0, 1)$.
For scale, we use exponential activation (which is $\mathbb{R} \xrightarrow{} \mathbb{R}^{+}$).
For opacity, we use the Sigmoid activation defined as $\sigma(x) = 1 / (1 + \exp(-x)) $ (which is $\mathbb{R} \xrightarrow{} (0, 1)$).
Besides the activations, we also want the initial output to be close to $0.1$. 
We accompolished this by adding constant bias to the transformer's output to shift the initialization.
We also clip the scales at a maximum size of $0.3$.
This max-scale clipping is applied because we empirically found that the scale of the 3D Gaussian can be very large (and the Gaussian will be degenerated to a long line) without such clipping. 
These line-shaped Gaussians can spread the gradients to multiple pixels after splatting and we found that it hurt the training stability.
In summary, according to the main method section, suppose $\mathrm{\mathbf{G}_{ij}}$ is the output parameters of Gaussians, and we split it into the components of  $(\mathbf{G}_\mathrm{rgb},\mathbf{G}_\mathrm{scale}, \mathbf{G}_\mathrm{rotation},  \mathbf{G}_\mathrm{opacity}, \mathbf{G}_\mathrm{distance})$; he equation for scale and opacity would be
\begin{align}
    \mathrm{scale} &= \min\{\exp{(\mathbf{G}_\mathrm{scale} - 2.3)}, 0.3\},  \\
    \mathrm{opacity} &= \sigma(\mathbf{G}_{\mathrm{opacity}} - 2.0),
\end{align}
where $\exp{(-2.3)}$ and $\sigma(-2.0)$ are both $0.1$ approximately. 
As the output $\mathrm{\mathbf{G}_{ij}}$ is designed to be zero-mean after the model-weight initialization (we also remove all bias terms as mentioned earlier). 
We can approximately get an output GS initialization that we desire.
Note that this initialization does not need to be accurate because it is mostly used to help training stability.

\boldstartspace{Rotation.} We predict unnormalized quaternions and use L2-normalize as activations to get unit quaternions. 

\boldstartspace{RGB.} We directly interpret the model output as the zero-order Spherical Harmonics coefficients used in Gaussian Splatting implementation~\cite{kerbl20233d}. We do not use higher-order Spherical Harmonics in this work for simplicity. We leave it to future work to improve the view-dependent modelling of our \ourmethod{}.

\boldstartspace{XYZ and Distance.}
In the main paper, we briefly mention that the Gaussian center is parameterized as $\mathrm{xyz} = \mathrm{ray}_o + t \cdot  \mathrm{ray}_d$.
To convert the transformers' output $\mathbf{G}_\mathrm{distance}$ to $t$, we employ an empirical near distance $d_\mathrm{near}$ and far distance $d_\mathrm{far}$. 
We then map our model output to the range of $(d_\mathrm{near}, d_\mathrm{far})$. 
The conversion is defined as 
\begin{align}
    \omega &= \sigma(\mathbf{G}_\mathrm{distance}), \\
    t &= (1 - \omega)\,\, d_\mathrm{near}  + \omega\,\, d_\mathrm{far}.
\end{align}
The $d_\mathrm{near}$ and $d_\mathrm{far}$ are set differently for the object-level \ourmethod{} and scene-level \ourmethod{}. 
For object-level \ourmethod{}, we use $d_\mathrm{near}=0.1$ and $d_\mathrm{far}=4.5$ and clip the predicted XYZ to be inside $[-1,1]^3$; this is aligned with our training data rendering setup.
For scene-level \ourmethod{}, we use $d_\mathrm{near}=0.0$ and $d_\mathrm{far}=500$ to account for the large depth variations of indoor and outdoor scenes.

\subsection{Camera Pose Normalization}
For object-level \ourmethod, we did not use any camera normalization, as the object sizes are pre-normalized before data generation. 
For scene-level \ourmethod{}, we first compute the mean camera pose by averaging all input camera poses and consider the coordinate frame of the mean camera as the world space, i.e. making input poses relative to the mean pose for subsequent processes. We then scale all input camera locations, so that they are located within the bounding box ($[-1,1]^3$) in the world space.